\documentclass[10pt,twocolumn,letterpaper]{article}

\usepackage[pagenumbers]{cvpr} % To force page numbers, e.g. for an arXiv version

\usepackage{graphicx}
\usepackage{amsmath}
\usepackage{amssymb}
\usepackage{booktabs}

\graphicspath{{pdf/}}
\usepackage{makecell}
\usepackage{multirow}
\usepackage{cite}
\usepackage{color}
\definecolor{Y}{RGB}{230,220,180}
\definecolor{B}{RGB}{172,189,215}
\usepackage{mathrsfs}

\usepackage[accsupp]{axessibility}  % Improves PDF readability for those with disabilities.
\usepackage{marvosym}

\usepackage[pagebackref,breaklinks,colorlinks]{hyperref}

\usepackage[capitalize]{cleveref}
\crefname{section}{Sec.}{Secs.}
\Crefname{section}{Section}{Sections}
\Crefname{table}{Table}{Tables}
\crefname{table}{Tab.}{Tabs.}

\def\cvprPaperID{10615} % *** Enter the CVPR Paper ID here
\def\confName{CVPR}
\def\confYear{2023}

\begin{document}

\title{SIEDOB: Semantic Image Editing by Disentangling Object and Background}

% \author{
% Wuyang Luo$^1$\orcidlink{0000-0002-1447-3445} \ \ \ \ \ \ \ \ 
% Su Yang$^1$\textsuperscript{\Letter} \ \ \ \ \ \ \ \ \ \ 
% Xinjian Zhang$^1$ \ \ \ \ \ \ \ \ \ \ 
% Weishan Zhang$^2$ \ \ \ \ \ \ \ \ \ \  \\
% % Zihang Lai$^3$\\
% % Victor Kulikov$^4$ \ \ \ \ \ \ \ \ \ \ 
% % Nikita Orlov$^4$\ \ \ \ \ \ \ \ \ \ 
% % Humphrey Shi$^{4, 5}$ \!\textsuperscript{\Envelope}\ \ \ \ \ \ \ \ \ \ 
% % Gao Huang$^{1}$ \!\textsuperscript{\Envelope}\\
% {\small$^1$Shanghai Key Laboratory of Intelligent Information Processing, School of Computer Science, Fudan University \\
% \small\email{\{wyluo18, suyang\}@fudan.edu.cn}} \\
% {\small$^2$School of Computer Science and Technology, China University of Petroleum}
% }
% \maketitle

% \let\thefootnote\relax\footnotetext{\textsuperscript{\Letter} Corresponding author}

\author{
Wuyang Luo$^1$ \ \ \ \ \ \ \ \ 
Su Yang$^1$\textsuperscript{\Letter} \ \ \ \ \ \ \ \ \ \ 
Xinjian Zhang$^1$ \ \ \ \ \ \ \ \ \ \ 
Weishan Zhang$^2$ \\
{\small$^1$Shanghai Key Laboratory of Intelligent Information Processing, School of Computer Science, Fudan University}\\
% {\small\email{\{wyluo18, suyang\}@fudan.edu.cn}}\\
{\tt\small \{wyluo18, suyang\}@fudan.edu.cn} \\
{\small$^2$School of Computer Science and Technology, China University of Petroleum}\\
}
\maketitle

\let\thefootnote\relax\footnotetext{\textsuperscript{\Letter} Corresponding author}

% \author{Wuyang Luo \and Su Yang \and Xinjian Zhang\\
% Shanghai Key Laboratory of Intelligent Information Processing, School of Computer Science, Fudan University\\
% \and
% \maketitle

%%%%%%%%%%%%%%%%%%%%%%%%%%%%%%%%%%%%%%%%%%%%%%%%%%%%%%%%%%%%%%%%%%%%%%%%%%%%%%%%%%%%%%%%%%%%%%%%%%%%%%%%%%%%%%%%%%%%%%%%%%%%%%%%%%%%%%%%%%%%%%%
%%%%%%%%% ABSTRACT
\begin{abstract}
  Semantic image editing provides users with a flexible tool to modify a given image guided by a corresponding segmentation map. In this task, the features of the foreground objects and the backgrounds are quite different. However, all previous methods handle backgrounds and objects as a whole using a monolithic model. Consequently, they remain limited in processing content-rich images and suffer from generating unrealistic objects and texture-inconsistent backgrounds. To address this issue, we propose a novel paradigm, \textbf{S}emantic \textbf{I}mage \textbf{E}diting by \textbf{D}isentangling \textbf{O}bject and \textbf{B}ackground (\textbf{SIEDOB}), the core idea of which is to explicitly leverages several heterogeneous subnetworks for objects and backgrounds. First, SIEDOB disassembles the edited input into background regions and instance-level objects. Then, we feed them into the dedicated generators. Finally, all synthesized parts are embedded in their original locations and utilize a fusion network to obtain a harmonized result. Moreover, to produce high-quality edited images, we propose some innovative designs, including Semantic-Aware Self-Propagation Module, Boundary-Anchored Patch Discriminator, and Style-Diversity Object Generator, and integrate them into SIEDOB. We conduct extensive experiments on Cityscapes and ADE20K-Room datasets and exhibit that our method remarkably outperforms the baselines, especially in synthesizing realistic and diverse objects and texture-consistent backgrounds. Code is available at \url{https://github.com/WuyangLuo/SIEDOB}.
\end{abstract}

%%%%%%%%%%%%%%%%%%%%%%%%%%%%%%%%%%%%%%%%%%%%%%%%%%%%%%%%%%%%%%%%%%%%%%%%%%%%%%%%%%%%%%%%%%%%%%%%%%%%%%%%%%%%%%%%%%%%%%%%%%%%%%%%%%%%%%%%%%%%%%%
%%%%%%%%% BODY TEXT
\section{Introduction}
Semantic image editing has recently gained significant traction due to its diverse applications, including adding, altering, or removing objects and controllably inpainting, outpainting, or repainting images. Existing methods have made impressive progress benefiting from Generative Adversarial Networks (GAN)~\cite{goodfellow2014generative,isola2017image} and have demonstrated promising results in relatively content-simple scenes, such as landscapes. However, they still suffer from inferior results for content-rich images with multiple discrepant objects, such as cityscapes or indoor rooms. This paper aims to improve editing performance in complex real-world scenes. 

\begin{figure}[t]
    \centering
    \includegraphics[width=8.5cm, trim=25 10 10 10,clip]{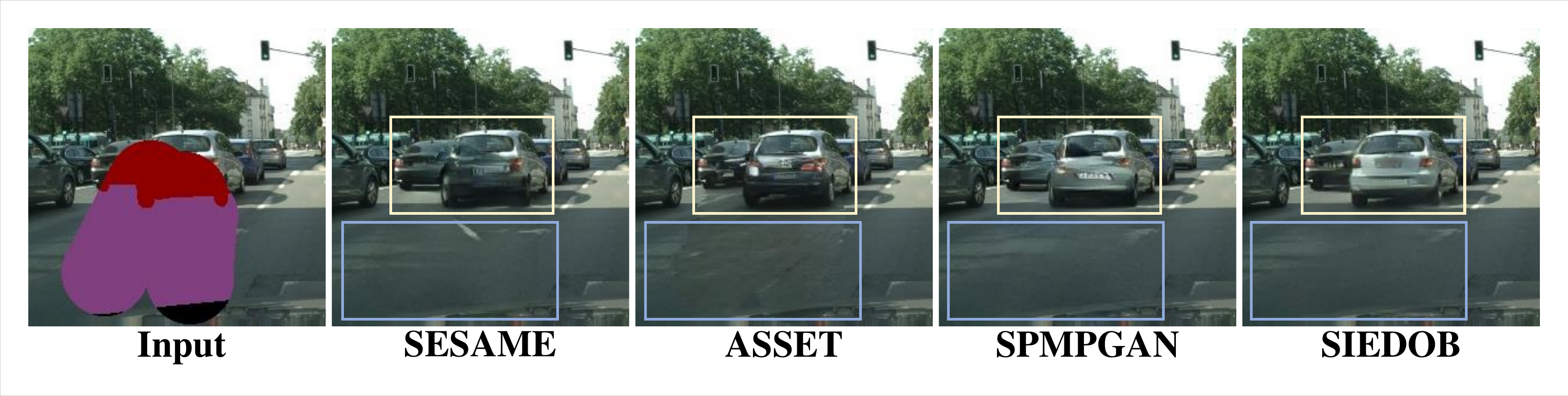}
    \caption{Existing methods struggle to deal with the compound of backgrounds and several overlapping objects in a complex scene. They generate distorted \textcolor{Y}{objects} and texturally inconsistent \textcolor{B}{backgrounds}. The proposed method can cope well with this input.}
    \label{fig:intro}
\end{figure}

When image editing experts deal with a complex task, they decompose the input image into multiple independent and disparate elements and handle them with different skills. For example, when editing a photo with a congested road, they process the cars and pedestrians individually and then fill in the background. Inspired by this spirit, we propose a heterogeneous editing model, which mimics the experience of human experts and explicitly splits the compound content into two distinct components: Backgrounds and objects. The background has no regular shape and may span a large area. More importantly, the re-generated background's texture and the existing area must be consistent. Foregrounds objects are class-specific and appear anywhere with various scales. Due to the significant differences between the two components, it is necessary to disentangle them from the input and process them with different networks. 

Several recent works \cite{SESAME,liu2022asset,luo2022context} have been dedicated to this task. SESAME \cite{SESAME} proposes a new pair of generator and discriminator based on the cGAN framework \cite{cGAN}, which generates edited content through a single-shot inference. ASSET \cite{liu2022asset} builds a transformer-based architecture with a sparse attention mechanism and synthesizes edited regions relying on a codebook generated by VQGAN \cite{esser2021taming}. SPMPGAN \cite{luo2022context} proposes a coarse-to-fine generator equipped with style-preserved modulation layers to retain style consistency between edited regions and context. These methods effectively enhance the ability of a monolithic model to handle the entire input. However, they deal with backgrounds and objects equally using the same generator, which causes them to produce inferior results. Figure \ref{fig:intro} demonstrates an example. To tackle this limitation, we present a novel framework,  \textbf{S}emantic \textbf{I}mage \textbf{E}diting by \textbf{D}isentangling \textbf{O}bject and \textbf{B}ackground (\emph{\textbf{SIEDOB}}), to generate backgrounds and foreground objects separately, which can achieve two goals: Synthesizing texture-consistent backgrounds and generating photo-realistic and diverse objects. 

\emph{SIEDOB} first disassembles the edited image into background regions and instance-level objects, then employs different generators to synthesize the corresponding content, and finally aggregates generated objects with backgrounds via a fusion network. In this way, we decouple our task into several more feasible subtasks and handle different components using dedicated sub-models. 

Generating texture-consistent backgrounds with known context is not trivial since backgrounds may cross a sizeable spatial area in an image and have no regular shape. We propose a pair of boosted generator and discriminator to address this problem. Specifically, we propose \emph{\textbf{S}emantic-\textbf{A}ware \textbf{S}elf-\textbf{P}ropagation \textbf{M}odule (SASPM)} to help the generator efficiently transfer the semantic-wise features of known regions to generated regions. Moreover, to further enhance texture consistency, we design a \emph{Boundary-Anchored Patch Discriminator} to force the generator to pay more attention to local textures of editing fringe. 

In our task, depending on the scenario, the edited object may be requested to be inpainted or generated. If an object is partially visible, it undergoes a lightweight inpainting network. Otherwise, we employ a \emph{Style-Diversity Object Generator} to obtain multi-modal results based on a style bank. After generating the backgrounds and all objects, we re-integrate them into a whole. However, the separate generation may lead to sudden boundaries and dissonance between objects and surroundings. To tackle this problem, we utilize a simple fusion network that predicts the residual value for each pixel to harmonize outputs. 

Our main contributions can be summarized as follows: 

\begin{itemize}
\item We propose a new solution for semantic image editing, named \emph{SIEDOB}, which can handle complex scenes by decoupling the generation of backgrounds and objects. 
\item We propose \emph{Semantic-Aware Self-Propagation Module} and \emph{Boundary-Anchored Patch Discriminator} to facilitate texture-consistent background generation. 
\item We propose a \emph{Style-Diversity Object Generator} that can generate diverse and realistic object images from masks.
\item Extensive experiments on two benchmarks indicate that our method can generate texture-consistent results and deal with crowded objects well.
\end{itemize} 

\begin{figure*}[t]
    \centering
    \includegraphics[width=16.0cm, trim=10 10 10 10,clip]{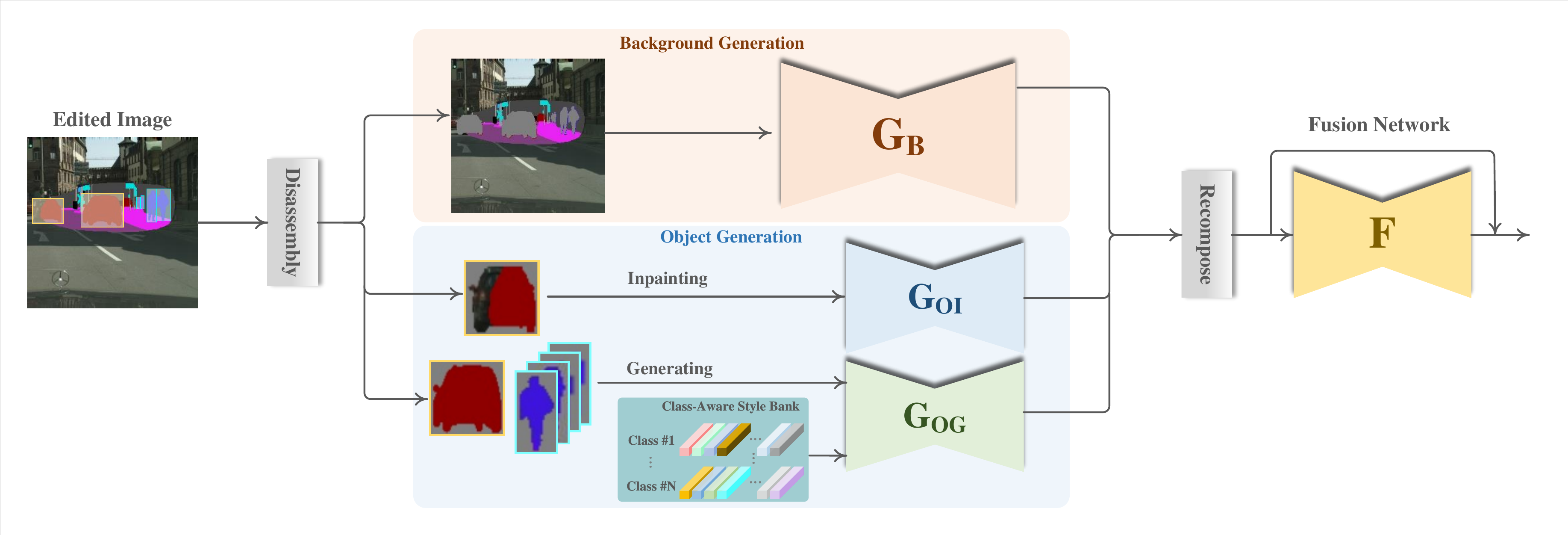}
    \caption{Overview of the proposed method. To reduce the complexity of modeling the entire edited region, we use a heterogeneous model to synthesize foreground objects and backgrounds separately, which contains three components: (1) Background generation, (2) object generation, (3) fusion network.}
    \label{fig:framework}
\end{figure*}

%%%%%%%%%%%%%%%%%%%%%%%%%%%%%%%%%%%%%%%%%%%%%%%%%%%%%%%%%%%%%%%%%%%%%%%%%%%%%%%%%%%%%%%%%%%%%%%%%%%%%%%%%%%%%%%%%%%%%%%%%%%%%%%%%%%%%%%%%%%%%%%
\section{Related Works}
\subsection{Semantic Image Editing}
Semantic image editing refers to modifying a given image by manipulating a corresponding segmentation map and keeping the context coherent. This approach provides powerful editing capabilities, such as adding and removing objects, modifying shapes, and re-painting backgrounds. Compared to several related tasks, including semantic image synthesis \cite{isola2017image,wang2018high,park2019semantic,tang2020local,zhu2020sean} and image inpainting \cite{iizuka2017globally,yu2018generative,yu2019free,zhao2020large,luo2023reference}, semantic image editing has not been fully exploited due to its difficulty in generating realistic objects and texture-consistent backgrounds simultaneously. A few recent works are devoted to semantic image editing \cite{HIM,SESAME,liu2022asset,luo2022context}. HIM \cite{HIM} is an early exploration that only operates on a single object using a two-stage network. SESAME \cite{SESAME} introduces a new pair of generators and discriminators to improve the quality of the generated results. Moreover, SESAME adopts a more flexible workflow that can respond to various image editing requirements of users. ASSET \cite{liu2022asset} proposes a novel transformer-based approach modeling long-range dependencies, enabling high-resolution image editing. SPMPGAN \cite{luo2022context} presents a style-preserved modulation technique and builds a progressive architecture that solves the style inconsistency problem in semantic image editing. These methods all consider semantic image editing as a global generation problem. Thus, they use a single model to process all elements of the edited image equally. In this work, we propose a decoupled framework to handle objects and backgrounds separately using heterogeneous generators, demonstrating more promising results. 

\subsection{Image Synthesis via Heterogeneous Generators}
For a wide variety of image generation tasks, most approaches impose monolithic generators on the entire image, including unsupervised image generation \cite{radford2015unsupervised,karras2019style,karras2020analyzing}, image-to-image translation \cite{wang2018high,park2019semantic,zhu2020sean, luo2022photo}, image inpainting \cite{yu2018generative,liu2018image,yu2019free}, and image editing \cite{liu2021deflocnet,jo2019sc}. They share the same network structure and weights to generate all the content without specialized submodels for different semantic regions or classes. Other methods use different architectures or weights to generate different image elements to improve generation quality for different components. For example, they employ multiple local branches specialized for different parts \cite{huang2017beyond,li2018global,gu2019mask,hinz2019generating}, decoupling content and style \cite{lee2018diverse,huang2018multimodal,liu2019few}, generating foreground and a single foreground object separately \cite{yang2017lr,singh2019finegan}, enhancing local generation \cite{shen2019towards,bhattacharjee2020dunit}, or having separate modules for different semantic categories \cite{tang2020local,li2021collaging}. However, these methods are developed for pure image generation or translation tasks where their input is noises, categories, segmentation maps, or full images. Therefore, they only need to synthesize simple but visually plausible results to fool the discriminator. Semantic image editing is a more challenging task, and its difficulty lies in keeping coherence between edited and known regions. In this paper, we propose a heterogeneous framework that utilizes different sub-models for foreground objects and backgrounds to improve their generation quality concurrently. 

%%%%%%%%%%%%%%%%%%%%%%%%%%%%%%%%%%%%%%%%%%%%%%%%%%%%%%%%%%%%%%%%%%%%%%%%%%%%%%%%%%%%%%%%%%%%%%%%%%%%%%%%%%%%%%%%%%%%%%%%%%%%%%%%%%%%%%%%%%%%%%%
\section{Proposed Method}
Our goal is to edit a given image $I \in \mathbb{R}^{H \times W \times 3}$ guided by a user-provided segmentation map $S \in \mathbb{L}^{H \times W \times L}$ within an edited region defined by a mask map $M \in \mathbb{B}^{H \times W \times 1}$ whose value is 0 in the non-edited region and 1 in the edited region. Here $H$, $W$, and $L$ denote the height, width, and number of semantic categories, respectively. The edited region of $I$ is erased as input ${I_{e}} = I \times(1-M)$. We create a background mask $M_B$ and all instance-level object masks $\left\{M_{c}^{q}\right\}$ in the edited region based on the segmentation map. Here, $c \in \mathcal{C}$, $\mathcal{C}$ is a pre-defined set of considered foreground categories, and $M_c^q$ is the $q$-th instance that falls into the category $c$. The background mask $M_B$ is composed of the remaining pixels. Thus, we can disassemble the input image $I_{e}$ to the background and objects according to $M_B$ and $\left\{M_{c}^{q}\right\}$. Then we generate the background and objects separately using the respective generators. Finally, we integrate all generated content via a fusion network. The workflow is depicted in Figure \ref{fig:framework}.

\subsection{Background Generation}
For image editing, the background usually needs to be partially generated to fill the space, such as sky or ground. Therefore, generating style-consistent texture patterns in edited regions is critical and challenging since the generated background and the known region coexist. In addition, an image's background may be across a distant spatial distance and have no regular shape. Efficiently and accurately transferring feature styles from known to edited regions is vital for generating texture-consistent results. To this end, we propose a \emph{\textbf{S}emantic-\textbf{A}ware \textbf{S}elf-\textbf{P}ropagation \textbf{M}odule (SASPM)}, which explicitly extracts semantic-wise feature codes from feature maps and propagates them to the corresponding regions at the same layer. Furthermore, we employ a \emph{Boundary-Anchored Patch Discriminator} to enforce the generator to focus on local textures at the editing boundary. 

\begin{figure}[t]
    \centering
    \includegraphics[width=7.5cm, trim=10 10 10 10,clip]{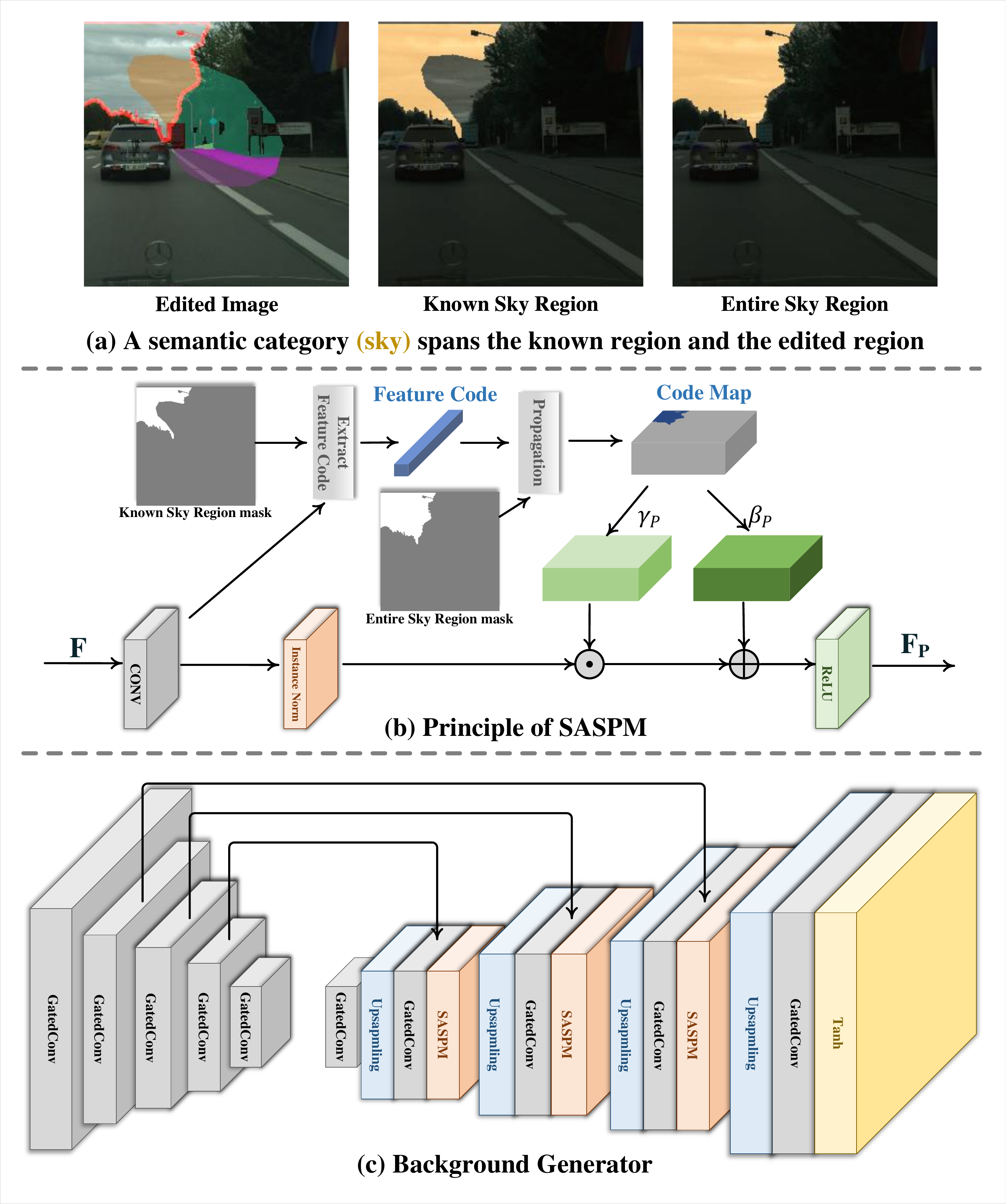}
    \caption{(a) A background category, such as sky, may span known and edited regions. We want to transfer the features from the known region to the generated region in a semantic-aware fashion. (b) The principle of \emph{SASPM}. We extract the feature code for each background category from the known region and then propagate it to the entire region. (c) The architecture of our background generator.}
    \label{fig:saspm}
\end{figure}

The proposed background generator $G_{B}$ is shown in \ref{fig:saspm}(c). 
We feed the disassembled background region into $G_{B}$ as the input: $I_B = {I_{e}} \times M_B$. $G_{B}$ follows an encoder-decoder structure. The encoder is composed of several successive GatedConv layers \cite{yu2019free} with stride two, and the decoder contains corresponding GatedConv layers with upsampling and several \emph{SASPM}. 

\noindent
\textbf{\emph{Semantic-Aware Self-Propagation Modules (SASPM)}}

\emph{SASPM} aims at propagating features of known regions to edited regions in a semantic-aware manner. Taking the sky area as an example shown in Figure \ref{fig:saspm}(a)(b), a \emph{SASPM} is divided into two steps. First, we extract the semantic-aware feature code by computing the average of the known sky area. Then, the feature code is broadcast into the entire sky area according to the corresponding semantic mask. \emph{SASPM} can transfer the feature codes of each semantic label at the same feature level without the constraint of the receptive field. Formally, let $F \in \mathbb{R}^{H \times W \times C}$ denote a feature map in front of a \emph{SASPM}. $H$, $W$ are the spatial dimension and $C$ is the number of channels. Let $U \in \mathbb{R}^{L \times C}$ represent $L$ feature codes extracted from the known region, and the feature codes of non-background classes are set to zero. We generate a code map $P$ based on the segmentation map $S \in \mathbb{L}^{H \times W \times L}$ through a matrix multiplication to spatially broadcast the feature codes into the corresponding semantic region: 

\begin{equation}
P = S \otimes U
\end{equation}

\noindent
Thus, $P \in \mathbb{R}^{H \times W \times C}$ is restored to the spatial dimension filled with the category-specific feature codes. Finally, we propagate $P$ into the original $F$ using a modulation operation following SPADE \cite{park2019semantic}. Specifically, we learn two parameters $\gamma_{P} \in \mathbb{R}^{H \times W \times C}$ and $\beta_{P} \in \mathbb{R}^{H \times W \times C}$ from $P$ to modulate $F$: 

\begin{equation}
F_{P} = ReLU\left(\gamma_{P} \odot IN({Conv(F)}) \oplus \beta_{P}\right)
\end{equation}

\noindent
Where $Conv(\cdot)$, $IN(\cdot)$, and $ReLU(\cdot)$ denote Convolutional Layer, Instance Normalization, and ReLU activation function, respectively. $\odot$ and $\oplus$ are element-wise multiplication and addition. 

\begin{figure}[t]
    \centering
    \includegraphics[width=7.3cm, trim=10 10 10 10,clip]{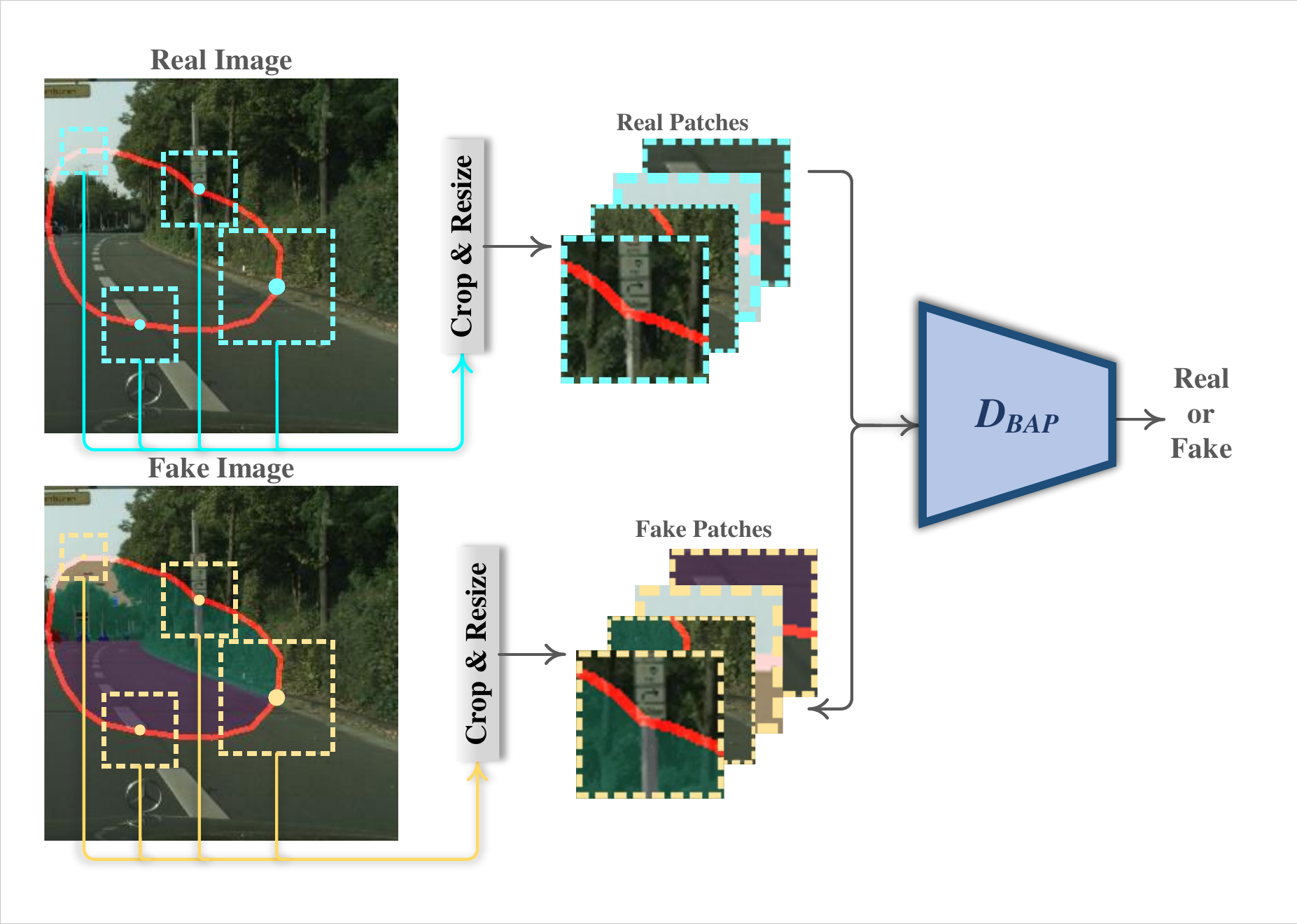}
    \caption{Boundary-Anchored Patch Discriminator ${D_{BAP}}$ takes in four real image patches or fake image patches of random size. All patches are centered on the boundary of the edited region enclosed by the red line.}
    \label{fig:Dpatch}
\end{figure}

\noindent
\textbf{\emph{Boundary-Anchored Patch Discriminator}}

We employ two discriminators for our background generator: (1) A global discriminator used by previous works \cite{luo2022context}; (2) a novel \emph{\textbf{B}oundary-\textbf{A}nchored \textbf{P}atch Discriminator} ${D_{BAP}}$. We introduce ${D_{BAP}}$ to enforce the generator to pay attention to local texture across the fringe of the edited area. We expect ${D_{BAP}}$ can capture known and generated textures simultaneously to facilitate texture-consistent results. To this end, we randomly select several central points anchored on the boundary of the edited region to crop patches from real or generated images, as shown in Figure \ref{fig:Dpatch}. Thus, the fake patches contain both real and fake textures, so ${D_{BAP}}$ can distinguish between true and false through texture consistency. 

The training objective of $G_{B}$ is comprised of \emph{L1 Distance Loss} $\mathcal{L}_{\mathrm{1}}$, \emph{Perceptual Loss} $\mathcal{L}_{\mathrm{P}}$ \cite{johnson2016perceptual},  \emph{Global Adversarial Loss} $\mathcal{L}_{GAN}^{G}$, and \emph{Local Patch Adversarial Loss} $\mathcal{L}_{GAN}^{L}$: 

\begin{equation}
\mathcal{L}_{B} = \mathcal{L}_{\mathrm{1}} + \lambda_{PB} \mathcal{L}_{\mathrm{P}} + \mathcal{L}_{GAN}^{G} + \lambda_{GAN}^{L} \mathcal{L}_{GAN}^{L}
\end{equation}

\noindent
Where $\mathcal{L}_{\mathrm{1}}$ and $\mathcal{L}_{\mathrm{P}}$ force the generated results to be closer to the ground truth in RGB space and VGG space \cite{vgg}. $\mathcal{L}_{GAN}^{G}$ and $\mathcal{L}_{GAN}^{L}$ are applied to global images and local patches, respectively. All adversarial losses in this work are associated with a SNPatchGAN Discriminator \cite{yu2019free} using the hinge version \cite{brock2018large}.

\begin{figure*}[t]
    \centering
    \includegraphics[width=15.5cm, trim=10 10 10 10,clip]{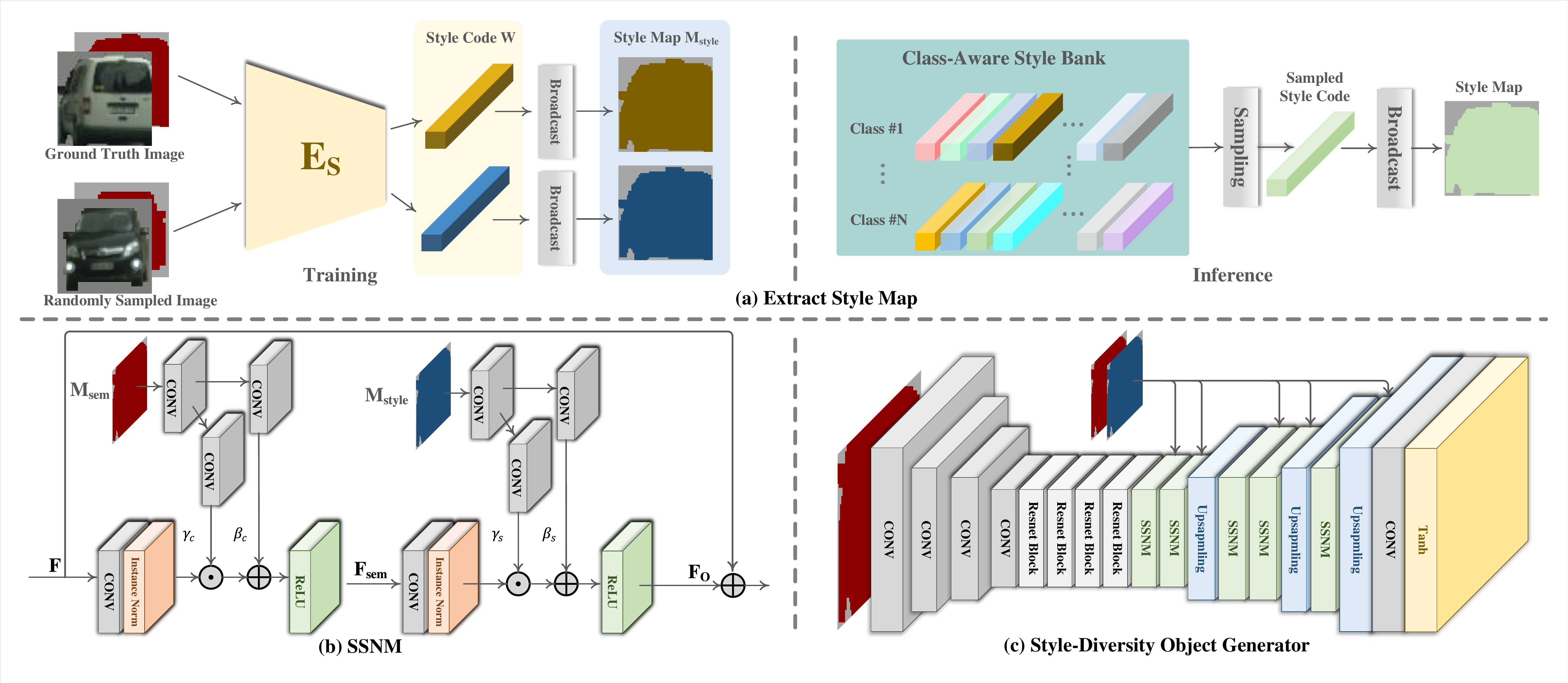}
    \caption{(a) How to extract style maps? During training, we extract their style codes from ground truth images or randomly sampled images and then generate style maps by broadcasting. During inference, we sample a class-specific style code directly from the style bank. (b) The structure of \emph{Semantic-Style Normalization Module (SSNM)}. (c) We build a \emph{Style-Diversity Object Generator} equipped with \emph{SSNM}.}
    \label{fig:object}
\end{figure*}

\subsection{Object Generation}
Our task involves dealing with two scenarios: Inpainting or generating an object—the former aims at completing an object based on its visible part. The latter synthesizes an inexistent object from its mask with an arbitrary appearance. Unlike handling all objects at once, cropping each edited object and independently generating them allows us to use aligned input images with the same size, which is beneficial for generating high-quality results. Specifically, we crop all objects $\left\{Crop(I_{e})_{c}^{q}\right\}$ from the edited region based on object masks $\left\{M_{c}^{q}\right\}$. We leverage a two-way model to handle the different scenarios. One way is a lightweight inpainting network  $G_{OI}$ to complete objects. 
$G_{OI}$ is a UNet-like \cite{unet} network and its training objective is: $\mathcal{L}_{GI}=\mathcal{L}_1+\lambda_{PI} \mathcal{L}_{\mathrm{P}}+\mathcal{L}_{GAN}$. $\mathcal{L}_1$, $\mathcal{L}_{\mathrm{P}}$, and $\mathcal{L}_{GAN}$ are the same as the terms in Equation 6.
The other way is the proposed \emph{Style-Diversity Object Generator} $G_{OG}$ for the multi-modal generation. Note that all categories of objects share $G_{OI}$ and $G_{OG}$.

\noindent
\textbf{\emph{Style-Diversity Object Generator}}

$G_{OG}$ generates a class-specific image from a one-hot segmentation map $M_{sem} \in \mathbb{L}^{H \times W \times K}$ created by the corresponding object mask $M_{c}^{q}$. $K$ is the number of the pre-defined foreground class set. Motivated by disentanglement learning \cite{lee2018diverse,huang2018multimodal,liu2019few}, we utilize style information to control the appearance of results for muti-modal generation. Specifically, we sample an image from the training set with the same class label and extract its style code $W \in \mathbb{R}^{128}$ by a style encoder $E_{s}$. Then we generate a style map $M_{style} \in \mathbb{R}^{H \times W \times 128}$ via broadcasting the style code according to $M_{c}^{q}$, as illustrated in Figure \ref{fig:object}(a). At this point, we have acquired a semantic map $M_{sem}$ and a style map $M_{style}$. The next is how to decode them back to a style-controlled object image. To this end, we design a \emph{Semantic-Style Normalization Module (SSNM)} to integrate the style and semantic information, as shown in Figure \ref{fig:object}(b). We first inject the semantic information into the input feature maps $F \in \mathbb{R}^{H \times W \times C}$:

\begin{equation}
F_{sem} = ReLU\left(\gamma_{c} \odot IN({Conv(F)}) \oplus \beta_{c}\right)
\end{equation}

\noindent
Where two normalization parameters, $\gamma_{c} \in \mathbb{R}^{H \times W \times C}$ and $\beta_{c} \in \mathbb{R}^{H \times W \times C}$ are learned from $M_{sem}$. Similarly, the style information is injected by:

\begin{equation}
F_{O} = ReLU\left(\gamma_{s} \odot IN({Conv(F_{sem})}) \oplus \beta_{s}\right)
\end{equation}

\noindent
Where two normalization parameters  $\gamma_{s} \in \mathbb{R}^{H \times W \times C}$ and $\beta_{s} \in \mathbb{R}^{H \times W \times C}$ are learned from $M_{style}$. 

During training, for each sample, we take the ground truth image $I^{gt}$ and a randomly sampled image $I^{s}$ as the style images to generate two results $R^{gt}$ and $R^{s}$. We employ the following objective function to optimize $G_{OG}$:

\begin{equation}
\mathcal{L}_{GO} = \mathcal{L}_{\mathrm{1}} + \lambda_{PO} \mathcal{L}_{\mathrm{P}} + \mathcal{L}_{GAN} + \mathcal{L}_{\mathrm{SCC}}
\end{equation}

\noindent
Here, $\mathcal{L}_{\mathrm{1}}$ is only applied to  $R^{gt}$. $\mathcal{L}_{\mathrm{P}}$, $\mathcal{L}_{GAN}$, and $\mathcal{L}_{\mathrm{SCC}}$ are imposed on all results. We introduce \emph{Style Cycle-Consistency Loss} $\mathcal{L}_{\mathrm{SCC}}$ to force the generator to produce style-consistent results with style images. 

% \begin{equation}
% \mathcal{L}_{SCC} = 1 - \boldsymbol\cos \left(\left\|E_s{ }^{\prime}(R)\right\|_2,\left\|E_s{ }^{\prime}({I}^{style})\right\|_2\right)
% \end{equation}

\begin{equation}
\mathcal{L}_{SCC} = 1 - \boldsymbol\cos \left(\frac{E_S^{\prime}(R)}{\left\|E_S^{\prime}(R)\right\|_2}, \frac{E_s^{\prime}\left(I^{style}\right)}{\left\|E_S^{\prime}\left(I^{style}\right)\right\|_2}\right)
\end{equation}

\noindent
Where $I^{style}$ is the style image and $R$ is the generated result. $E_s^{\prime}(\cdot)$ is a style encoder with the same structure as $E_s$. 
% $\|\cdot\|_2$ is $L2$ normalization. 
$\cos (\cdot, \cdot)$ represents the cosine similarity of two vectors.

Previous image translation methods \cite{lee2018diverse,huang2018multimodal} achieve multi-modal generation by adding Gaussian noise during inference. We find that noise leads to low-quality results and limited diversity. To solve this problem, we replace the noise with a style bank. Precisely, once training is done, we extract all style codes from the training set by the trained $E_{s}$ and save them for building a 
class-aware style bank. During inference, we randomly sample different style codes from the style bank to generate different results. It is worth mentioning that our approach does not increase any additional computational overhead compared to adding noise.

\begin{figure}[t]
    \centering
    \includegraphics[width=8.0cm, trim=10 10 10 10,clip]{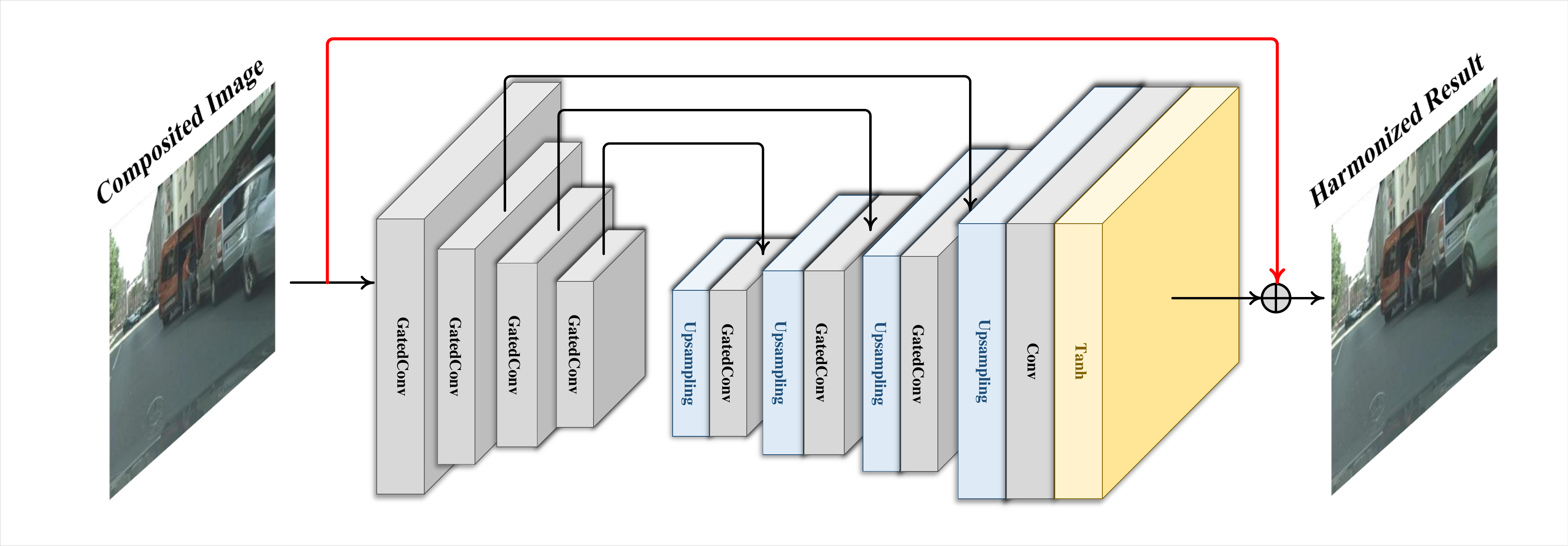}
    \caption{Structure of Fusion Network. The red line indicates the skip connection.}
    \label{fig:fusion_net}
\end{figure}

\subsection{Fusion Network}
After generating all backgrounds and objects, we embed them into their original positions to obtain a composited image. However, this preliminary result may be discordant since each object is generated independently without accounting for context. We employ a fusion network ${F}$ to harmonize the objects with the surroundings, as shown in Figure \ref{fig:fusion_net}. ${F}$ learns an offset value for each pixel via a skip connection \cite{unet,he2016deep} to produce the final results. We use \emph{Perceptual Loss} and \emph{Adversarial Loss} to optimize ${F}$: 

\begin{equation}
\mathcal{L}_{F} = \lambda_{PF} \mathcal{L}_{\mathrm{P}} + \mathcal{L}_{GAN}
\end{equation}

%%%%%%%%%%%%%%%%%%%%%%%%%%%%%%%%%%%%%%%%%%%%%%%%%%%%%%%%%%%%%%%%%%%%%%%%%%%%%%%%%%%%%%%%%%%%%%%%%%%%%%%%%%%%%%%%%%%%%%%%%%%%%%%%%%%%%%%%%%%%%%%

\section{Experiments}
\noindent
\textbf{Datasets.} We conduct experiments on two datasets, Cityscapes and ADE20K-Room, used in \cite{SESAME,luo2022context}. Cityscapes \cite{cordts2016cityscapes} contains complex street scene images in German cities. ADE20K-Room is a subset of ADE20K \cite{zhou2017scene} consisting of indoor scene images. We train and test our model at $256 \times 256$ resolution. 

\noindent
\textbf{Implementation Details.} We use free-form, extension, and outpainting masks following \cite{luo2022context}. Similar to the previous work \cite{li2021collaging}, we choose those frequently-appearing categories as the foreground class set, (car, person) for Cityscapes and (bed, chest, lamp, chair, table) for ADE20K-Room. All sub-networks are independently trained using ADAM optimizers \cite{adam} for both the generator and the discriminators with momentum $\beta_{1}=0.5$ and $\beta_{2}=0.999$, and the learning rates for the generator and the discriminators are set to 0.0001 and 0.0004, respectively. For training ${D_{BAP}}$, we randomly crop 4 square patches with the size ranging from $96 \times 96$ to $160 \times 160$. We set balance coefficients: $\lambda_{PB} = \lambda_{PI} = \lambda_{PO} = \lambda_{PF} = 10$ and $\lambda_{GAN}^{L} = 0.2$. The proposed method is implemented with Pytorch 1.10 \cite{paszke2019pytorch}, and all experiments are performed on a single NVIDIA RTX 3090 GPU. 

\noindent
\textbf{Baselines.} We compare the proposed methods with the state-of-the-art semantic image editing methods, including HIM \cite{HIM}, SESAME \cite{SESAME}, ASSET \cite{liu2022asset}, SPMPGAN \cite{luo2022context}. We also introduce two additional methods \cite{tang2020local,zhao2020large} as baselines to investigate related work more widely. LGGAN \cite{tang2020local} learns a map from segmentation maps to photo-realistic images using global image-level and class-specific generators. Co-Mod \cite{zhao2020large} demonstrates impressive results in semantic image generation and image inpainting. We change their input to be the same as our task. 

\begin{figure}[t]
    \centering
    \includegraphics[width=8.5cm, trim=10 10 10 10,clip]{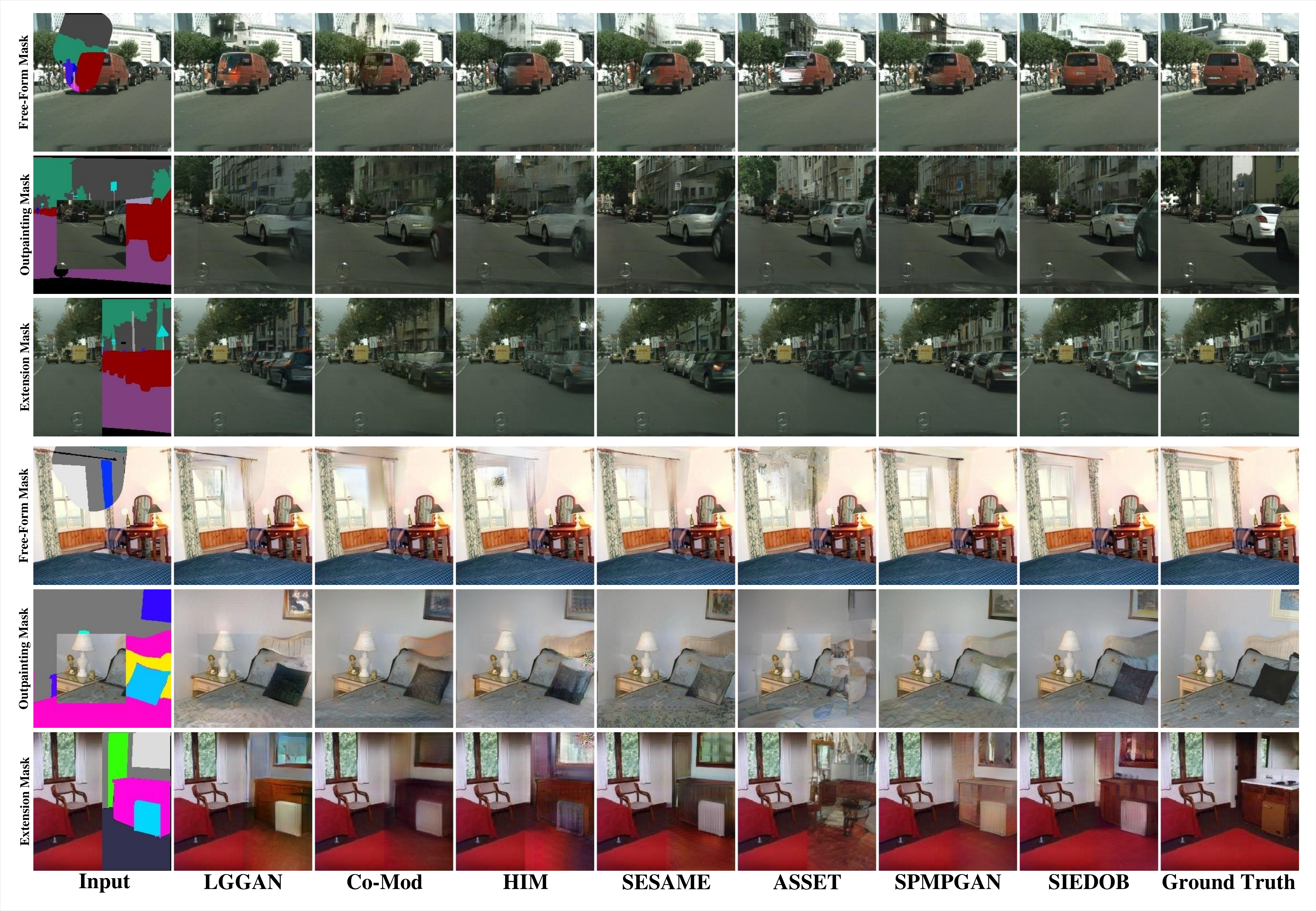}
    \caption{Visual comparisons with state-of-the-art methods. Our method is effective in generating texture-consistent backgrounds and photo-realistic objects.}
    \label{fig:sota}
\end{figure}

\subsection{Qualitative evaluation} 
We demonstrate a visual comparison of the competing methods, as shown in Figure \ref{fig:sota}. We observe that the results produced by the proposed \emph{SIEDOB} are visually better than existing methods in two aspects: (1) Our method can well handle dense objects overlapping each other, e.g., in the second and third rows. However, other methods produce distorted and inseparable results. (2) Our method generate texture-consistent content between edited regions and known regions both on background and objects, e.g., the red cars in the first row and the curtain in the fourth row. The performance improvement benefits from our disentangling generation strategy and carefully designed sub-networks. 

\begin{table}[t]
\centering
\scriptsize
% \footnotesize
\setlength{\tabcolsep}{1.0mm}
\begin{tabular}{clcccccc}
\Xhline{1pt}
\multirow{2}{*}{M.} & \multirow{2}{*}{Method} & \multicolumn{3}{c}{Cityscapes} & \multicolumn{3}{c}{ADE20K-Room} \\ \cline{3-8} 
                    &                         & FID$\downarrow$      & LPIPS$\downarrow$    & mIoU$\uparrow$     & FID$\downarrow$       & LPIPS$\downarrow$    & mIoU$\uparrow$     \\ \Xhline{1pt}
\multirow{7}{*}{F.} & LGGAN                   & 15.29    & 0.094    & 58.62    & 24.74     & 0.097    & 27.68    \\
                    & Co-Mod                  & 15.88    & 0.097    & 56.50    & 27.37     & 0.111    & 27.52    \\
                    & HIM                     & 15.58    & 0.093    & 58.99    & 28.64     & 0.133    & 28.04    \\
                    & SESAME                  & 12.89    & 0.082    & 58.88    & 21.73     & 0.101    & 27.50    \\
                    & ASSET                   & 13.67    & 0.098    & 58.12    & 30.63     & 0.126    & 26.02    \\
                    & SPMPGAN                 & 11.90    & 0.084    & 58.80    & 18.83     & 0.090    & 28.22    \\
                    & SIEDOB                  & \textbf{11.07}    & \textbf{0.077}    & \textbf{59.41}    & \textbf{17.61}     & \textbf{0.089}    & \textbf{29.72}    \\ \Xhline{0.75pt}
\multirow{7}{*}{E.} & LGGAN                   & 26.01    & 0.175    & 58.37    & 37.09     & 0.213    & 27.91    \\
                    & Co-Mod                  & 29.27    & 0.188    & 56.44    & 38.61     & 0.231    & 27.13    \\
                    & HIM                     & 25.20    & 0.180    & 58.91    & 40.69     & 0.239    & 27.61    \\
                    & SESAME                  & 20.30    & 0.168    & 59.08    & 36.43     & 0.211    & 27.62    \\
                    & ASSET                   & 21.99    & 0.186    & 58.01    & 38.17     & 0.261    & 27.03    \\
                    & SPMPGAN                 & 19.46    & 0.167    & 59.10    & 32.92     & 0.199    & 27.73    \\
                    & SIEDOB                  & \textbf{19.20}    & \textbf{0.159}    & \textbf{59.63}    & \textbf{31.66}     & \textbf{0.191}    & \textbf{29.52}    \\ \Xhline{0.75pt}
\multirow{7}{*}{O.} & LGGAN                   & 39.12    & 0.254    & 58.77    & 49.07     & 0.334    & 27.69    \\
                    & Co-Mod                  & 50.29    & 0.264    & 55.39    & 51.45     & 0.325    & 26.54    \\
                    & HIM                     & 36.27    & 0.252    & 58.99    & 54.51     & 0.337    & 28.19    \\
                    & SESAME                  & 28.27    & 0.237    & 58.75    & 47.72     & 0.305    & 27.40    \\
                    & ASSET                   & 30.60    & 0.240    & 58.91    & 57.28     & 0.331    & 27.18    \\
                    & SPMPGAN                 & \textbf{27.63}    & 0.233    & 58.53    & 41.52     & 0.288    & 27.85    \\
                    & SIEDOB                  & 27.90    & \textbf{0.221}    & \textbf{59.37}    & \textbf{41.07}     & \textbf{0.275}    & \textbf{28.94}    \\ \bottomrule
\end{tabular}
\caption{Quantitative comparison with other methods. M., F., E., and O. represent Mask Type, Free-Form Mask, Extension Mask, and Outpainting Mask, respectively. ($\uparrow$: Higher is better; $\downarrow$: Lower is better)}
\label{tab:sota}
\end{table}

\subsection{Quantitative evaluation}
Table \ref{tab:sota} lists the quantitative results with three different mask types. Following previous works, we use three metrics: FID \cite{FID}, LPIPS \cite{LPIPS}, and mean Intersection-over-Union (mIoU). FID is introduced to assess the fidelity of the results by computing the Wasserstein-2 distance between the distributions of the synthesized and real images. LPIPS evaluates the similarity between the generated image and the ground truth in a pairwise manner. mIoU measures the alignment of semantic labels between the generated results and input segmentation maps. We use solid models to obtain segmentation maps: HRNet \cite{wang2020deep} for ADE20K-Room, and DRN-D-105 \cite{yu2017dilated} for Cityscapes. Our method outperforms the other methods under different metrics and mask settings. 

\begin{figure}[t]
    \centering
    \includegraphics[width=8.5cm, trim=10 10 10 10,clip]{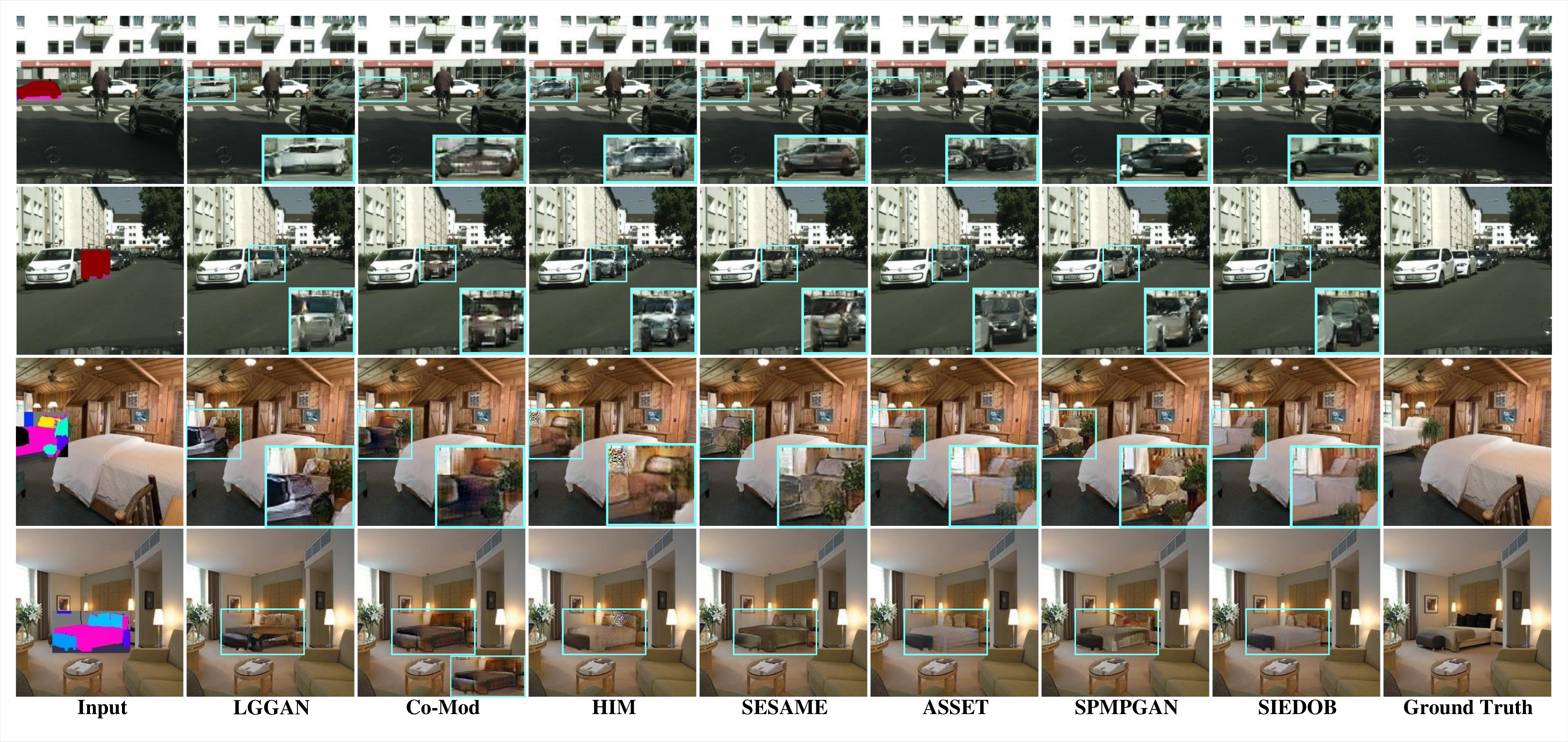}
    \caption{Visual results of addition objects.}
    \label{fig:addition}
\end{figure}

\begin{figure}[t]
    \centering
    \includegraphics[width=8.1cm, trim=10 10 10 10,clip]{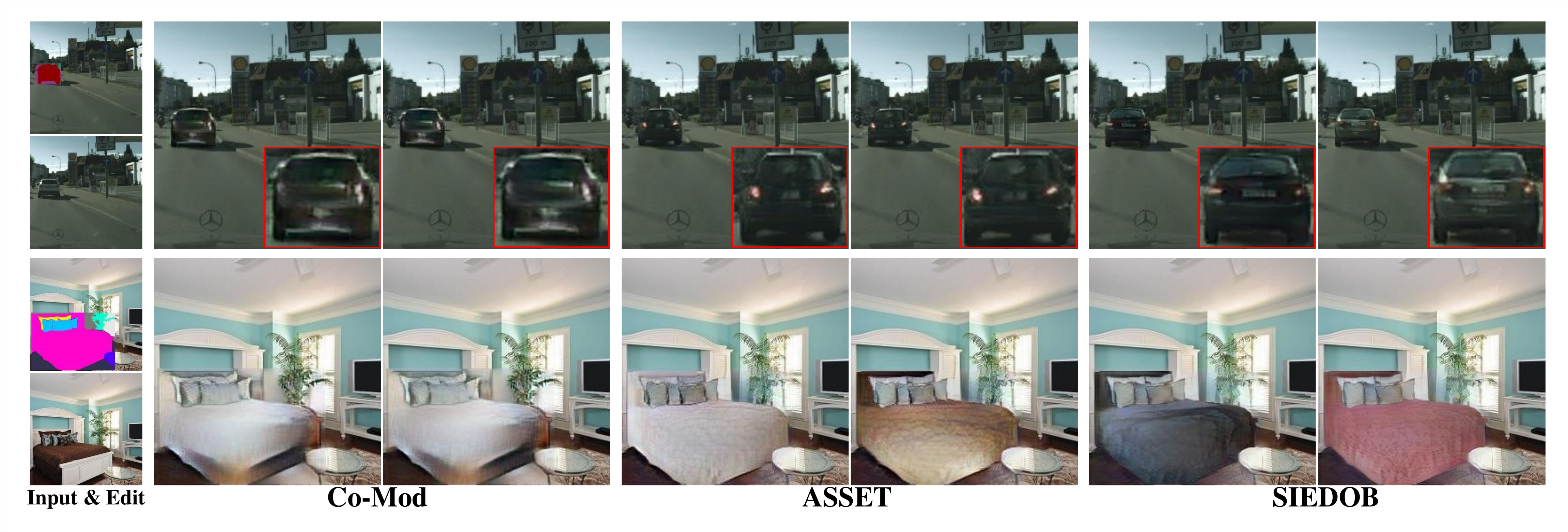}
    \caption{Multi-modal generation results.}
    \label{fig:div}
\end{figure}

\begin{figure}[t]
    \centering
    \includegraphics[width=7.1cm, trim=10 10 10 10,clip]{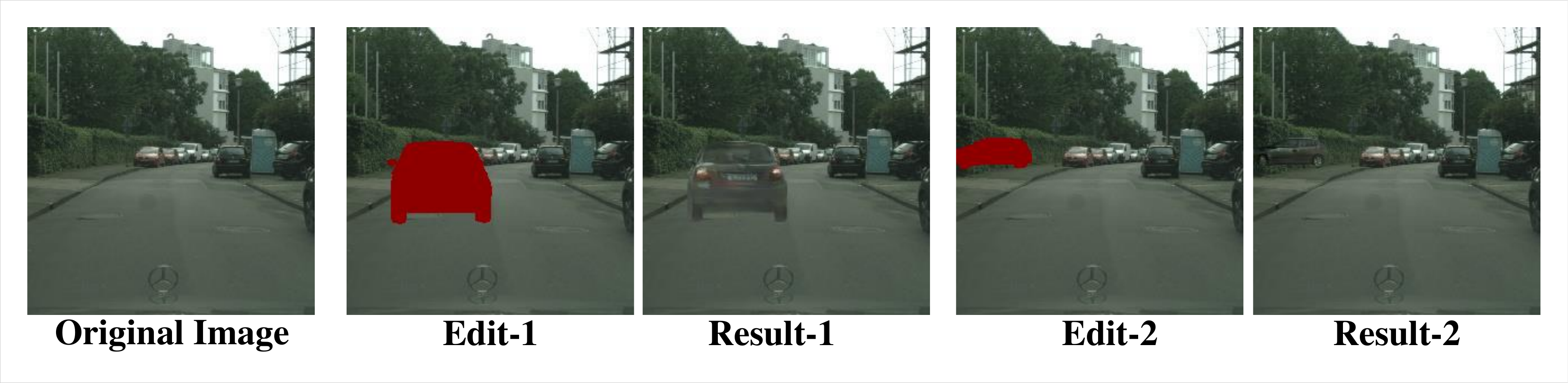}
    \caption{Out-of-distribution editing. Our method can add cars in the middle of the road or on the sidewalk, which do not exist in the training dataset.}
    \label{fig:out_dis}
\end{figure}

\subsection{Addition Object} 
Adding a new object to the given image is an essential ability of semantic image editing. The visual results are shown in Figure \ref{fig:addition}. In this experiment, we randomly select an instance of each test image and create a rectangular mask encircling the object. The proposed method can generate realistic objects with crisp edges.
Adding an object to the edited area that does not exist originally should have diverse results. Some previous methods \cite{liu2022asset, zhao2020large} have multi-modal generation capabilities. We sample two results from all methods, as shown in Figure \ref{fig:div}. 
Our method achieves a higher diversity than other models because the previous methods only exert a global influence on the generation process; in contrast, our model independently generates each object using different class-aware style codes. The quantitative results listed in Table \ref{tab:add} also indicate the superiority of our method. 
Here, the Diversity score (Div) is calculated as the average LPIPS distance between pairs of randomly sampled results for all test images, as also done in \cite{liu2021pd,zheng2019pluralistic}. Moreover, our model can hallucinate images that do not exist in the dataset's distribution, as shown in Figure \ref{fig:out_dis}. 

\begin{table}[]
\centering
\scriptsize
\setlength{\tabcolsep}{1.0mm}
\begin{tabular}{lccccccccc}
\Xhline{1pt}
\multirow{2}{*}{} & \multicolumn{4}{c}{Cityscapes} & \multicolumn{4}{c}{ADE20K-Room} \\ \cline{2-9} 
                  & FID$\downarrow$      & LPIPS$\downarrow$    & mIoU$\uparrow$    & Div$\uparrow$     & FID$\downarrow$       & LPIPS$\downarrow$    & mIoU$\uparrow$     & Div$\uparrow$     \\ \Xhline{1pt}
LGGAN             & 7.69     & 0.027    & 57.01     & -    & 24.49     & 0.171    & 30.05     & -   \\ \hline
Co-Mod            & 7.79     & 0.029    & 56.29     & 0.0030    & 21.09     & 0.118    & 30.63     & 0.0145   \\ \hline
HIM               & 6.92     & 0.028    & 58.71     & -    & 29.10     & 0.232    & 30.58     & -   \\ \hline
SESAME            & 6.67     & 0.024    & 58.79     & -   & 18.01     & 0.117    & 30.89     & -   \\ \hline
ASSET             & 6.74     & 0.025    & 56.02     & 0.0052   & 27.68     & 0.209    & 29.79     &  0.0337   \\ \hline
SPMPGAN           & 6.59     & 0.021    & 58.60     & -   & 17.21     & 0.115    & 30.58     & -   \\ \hline
SIEDOB            & \textbf{6.29}     & \textbf{0.020}    & \textbf{58.88}     & \textbf{0.0055}   & \textbf{16.32}     & \textbf{0.110}    & \textbf{31.04}     & \textbf{0.0362}   \\ \bottomrule
\end{tabular}
\caption{Quantitative results of object addition.}
\label{tab:add}
\end{table}

\begin{table}[t]
\centering
\scriptsize
\setlength{\tabcolsep}{1.0mm}
\begin{tabular}{lcccccc}
\Xhline{1pt}
\multirow{2}{*}{} & \multicolumn{3}{c}{Cityscapes} & \multicolumn{3}{c}{ADE20K-Room} \\ \cline{2-7} 
                  & FID$\downarrow$      & LPIPS$\downarrow$    & mIoU$\uparrow$     & FID$\downarrow$       & LPIPS$\downarrow$    & mIoU$\uparrow$     \\ \hline
\emph{w/o SASPM}         & 14.11    & 0.085    & 59.43    & 20.03     & 0.091    & 29.02    \\ \hline
\emph{w/o ${D_{BAP}}$}            & 12.68    & 0.081    & 59.49    & 19.14     & 0.083    & 29.21    \\ \hline
Full model        & \textbf{12.15}    & \textbf{0.079}    & \textbf{59.51}    & \textbf{18.72}     & \textbf{0.081}    & \textbf{29.30}    \\ \bottomrule
\end{tabular}
\caption{Quantitative study on \emph{SASPM} and ${D_{BAP}}$.}
\label{tab:saspm}
\end{table}

\subsection{Ablation Study} 

\begin{figure}[t]
    \centering
    \includegraphics[width=6.5cm, trim=10 10 10 10,clip]{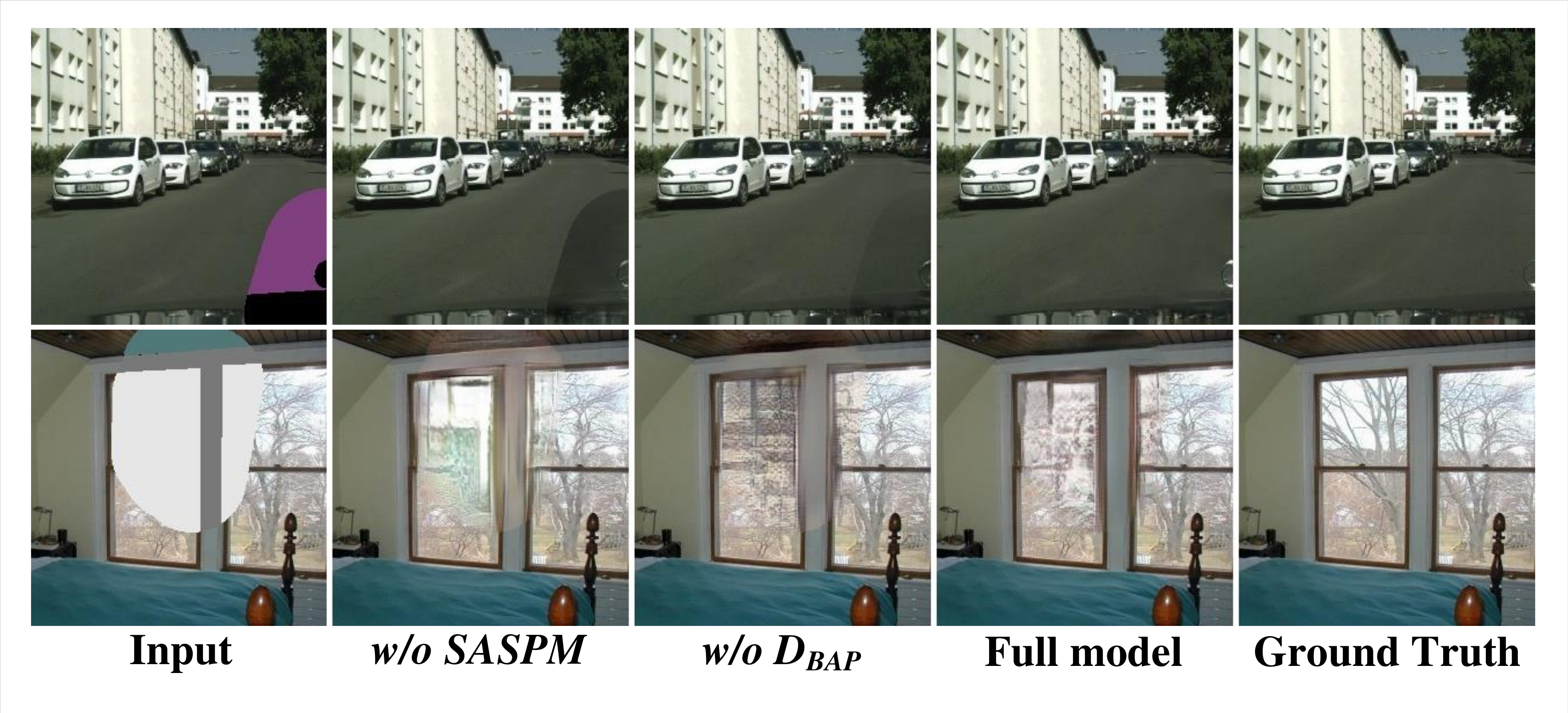}
    \caption{Ablation analysis on \emph{SASPM} and ${D_{BAP}}$ for synthesizing texture-consistent backgrounds.}
    \label{fig:bg}
\end{figure}

\noindent
\textbf{Effectiveness of \emph{SASPM} and \emph{Boundary-Anchored Patch Discriminator}.} We investigate the contributions of the proposed \emph{SASPM} and ${D_{BAP}}$ to the generation of texture-consistent background. To exclude the interference of the foreground, we create mask maps that only include background regions in this study. We remove all \emph{SASPM} ("\emph{w/o SASPM}") or ${D_{BAP}}$ ("\emph{w/o} ${D_{BAP}}$") from our model as two baselines. The visual results are shown in Figure \ref{fig:bg} and prove that \emph{SASPM} and ${D_{BAP}}$ can help the background generator effectively alleviate texture inconsistencies. Vanilla convolutional layers have limited receptive fields and no ability to locate where edited or known regions are. \emph{SASPM} explicitly transfers features from known regions to edited regions with the same semantic labels across distant spatial distances. ${D_{BAP}}$ forces the generator to pay more attention to local textures on editing boundaries rather than just global images. Quantitative results are listed in Table \ref{tab:saspm}.

\begin{figure}[t]
    \centering
    \includegraphics[width=8.1cm, trim=10 10 10 10,clip]{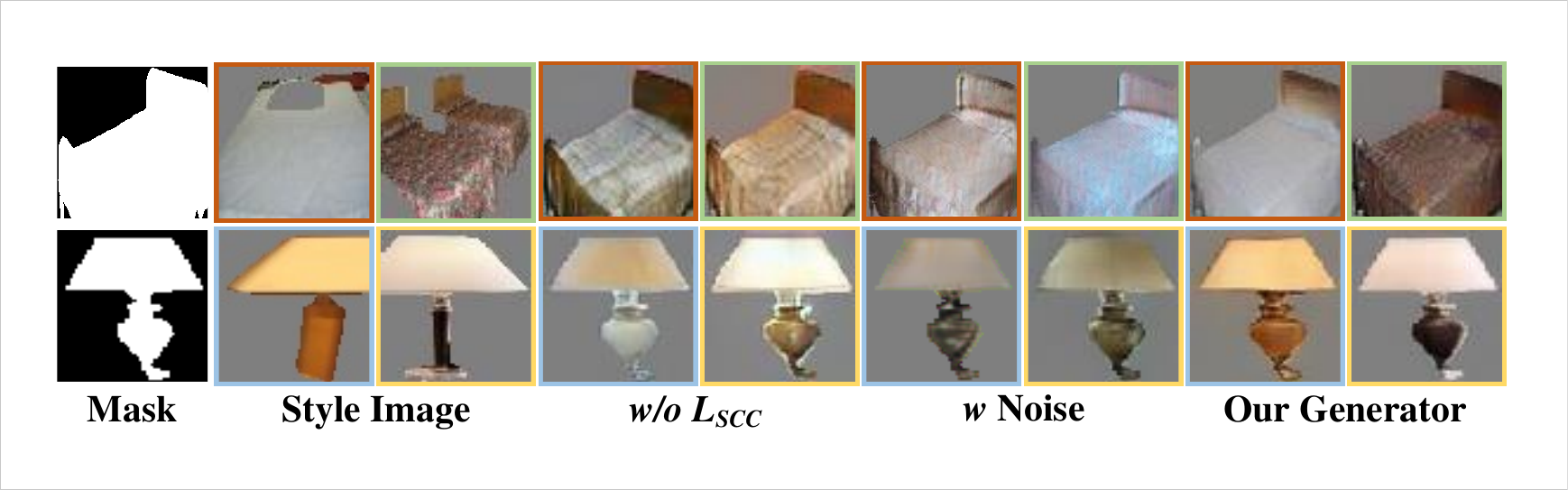}
    \caption{Ablation study of object generation.}
    \label{fig:ablation_object}
\end{figure}

\noindent
\textbf{Effectiveness of Style-Diversity Object Generator.} 
To verify the benefit of $\mathcal{L}_{\mathrm{SCC}}$ and the style bank, we set up two baselines: (1) Dropping out $\mathcal{L}_{\mathrm{SCC}}$ ("\emph{w/o} $\mathcal{L}_{\mathrm{SCC}}$"). (2) Replacing the style codes with noise vectors ("\emph{w} Noise"). Figure \ref{fig:ablation_object} shows some visual results. If we drop out $\mathcal{L}_{\mathrm{SCC}}$, the generated image's style may not be consistent with the style image. If we produce multi-modal results by injecting noise, it leads to a loss of fidelity and diversity. 
This is because Gaussian noise is semantics-free, while style code is category-specific and provides controllability.
 The quantitative results are listed in Table \ref{tab:obj}.

% Here, LPIPS is calculated between two generated results using different style codes. Larger LPIPS means the generated results are more diverse. 

\begin{figure}[t]
    \centering
    \includegraphics[width=7.4cm, trim=10 10 10 10,clip]{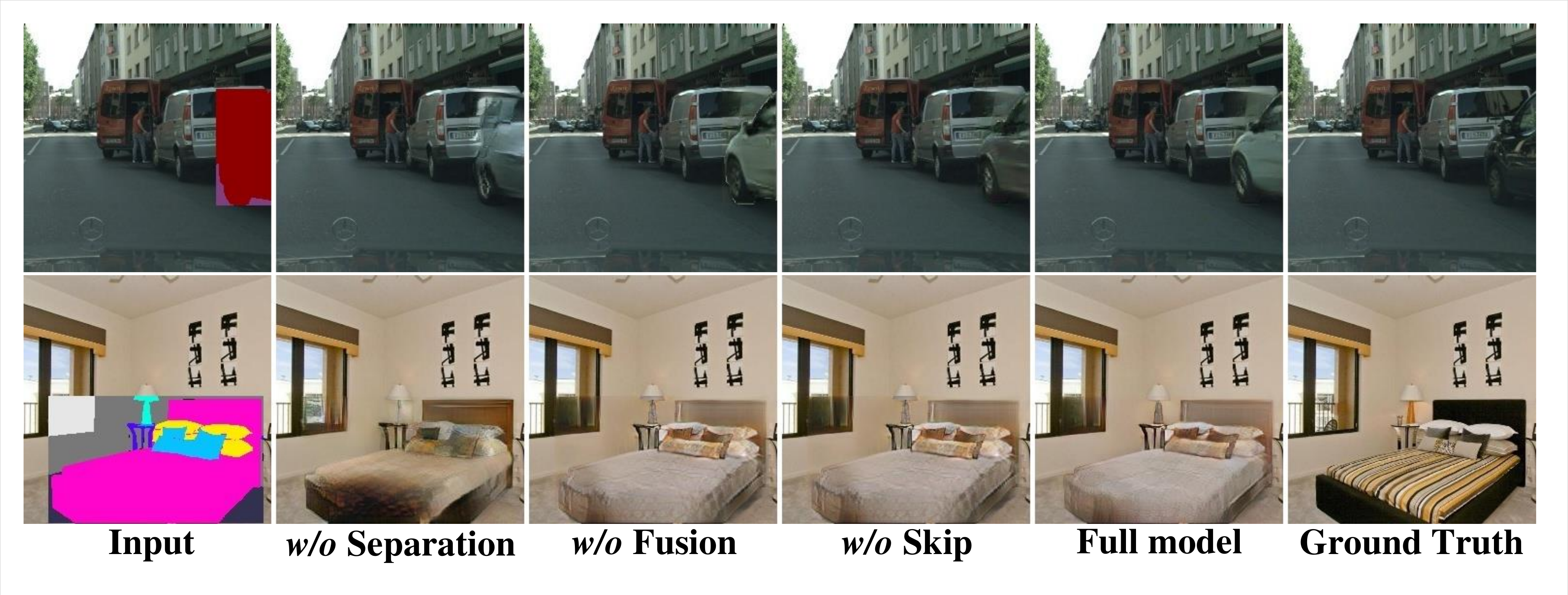}
    \caption{Visual analysis on disentangled generation and fusion network.}
    \label{fig:ablation}
\end{figure}

\noindent
\textbf{Effectiveness of Separate Generation.} We remove the object generator and the fusion network but keep other components unchanged to obtain a baseline (\emph{w/o} Separation) and increase its channel number for a fair comparison. Qualitative and quantitative comparisons can be found in Figure \ref{fig:ablation} and Table \ref{tab:fusion}, respectively.  In this experiment, we employ the masks used in object addition. The visual comparison shows that the separate generation paradigm significantly improves performance for objects with complicated surroundings. 

\begin{table}[t]
\centering
\scriptsize
\setlength{\tabcolsep}{1.0mm}
\begin{tabular}{lcccccc}
\Xhline{1pt}
\multirow{2}{*}{class} & \multicolumn{2}{c}{w/o $\mathcal{L}_{\mathrm{SCC}}$}                           & \multicolumn{2}{c}{w Noise}                            & \multicolumn{2}{c}{$G_{OG}$}                                \\ \cline{2-7} 
                       & FID$\downarrow$                        & Div$\uparrow$                       & FID$\downarrow$                        & Div$\uparrow$                       & FID$\downarrow$                        & Div$\uparrow$                       \\ \hline
car                    & 123.82                     & 0.103                     & 173.57                     & 0.090                     & \textbf{118.32}                     & \textbf{0.115}                     \\ \hline
person                 & 238.10                     & 0.079                     & 307.11                     & 0.052                     & \textbf{219.20}                     & \textbf{0.098}                     \\ \hline
bed                    & 227.18                     & 0.303                     & 395.33                     & 0.210                     & \textbf{239.78}                     & \textbf{0.372}                     \\ \hline
chest                  & 192.12                     & 0.217                     & 273.10                     & 0.113                     & \textbf{149.29}                     & \textbf{0.229}                     \\ \hline
lamp                   & 121.01                     & 0.195                     & 153.94                     & 0.129                     & \textbf{112.50}                     & \textbf{0.218}                     \\ \hline
chair                  & 168.01                     & 0.274                     & 207.35                     & 0.177                     & \textbf{163.14}                     & \textbf{0.289}                     \\ \hline
table                  & 211.39                     & 0.141                     & 289.55                     & 0.091                     & \textbf{201.32}                     & \textbf{0.143}                     \\ \bottomrule
\end{tabular}
\caption{Quantitative results of different objects.}
\label{tab:obj}
\end{table}

\begin{table}[t]
\centering
\scriptsize
\setlength{\tabcolsep}{1.0mm}
\begin{tabular}{lcccccc}
\Xhline{1pt}
\multicolumn{1}{c}{\multirow{2}{*}{}} & \multicolumn{3}{c}{Cityscapes} & \multicolumn{3}{c}{ADE20K-Room} \\ \cline{2-7} 
\multicolumn{1}{c}{}                  & FID$\downarrow$      & LPIPS$\downarrow$    & mIoU$\uparrow$     & FID$\downarrow$       & LPIPS$\downarrow$    & mIoU$\uparrow$     \\ \hline
\emph{w/o} Separation                       & 6.78     & 0.029    & 56.43    & 21.97     & 0.180    & 30.32    \\ \hline
\emph{w/o} Fusion                            & 6.73     & 0.027    & 58.68    & 19.19     & 0.131    & 30.80    \\ \hline
\emph{w/o} Skip                              & 6.41     & 0.022    & 58.76    & 17.89     & 0.118    & 30.96    \\ \hline
Full model                          & \textbf{6.29}     & \textbf{0.020}    & \textbf{58.88}   & \textbf{16.32}     & \textbf{0.110}    & \textbf{31.04}    \\ \bottomrule
\end{tabular}
\caption{Quantitative studies of separate generation paradigm and fusion network.}
\label{tab:fusion}
\end{table}

\noindent
\textbf{Effectiveness of Fusion Network.} Fusion network can eliminate abrupt boundaries between generated objects and backgrounds. Removing the fusion network (\emph{w/o} Fusion) or not using the skip connection (\emph{w/o} Skip) will degrade the performance, which are qualitatively and quantitatively demonstrated in Figure \ref{fig:ablation} and Table \ref{tab:fusion}.

\section{Conclusion and Limitations}
This paper proposes a novel paradigm for semantic image editing, named \emph{SIEDOB}. Its core idea is that since the characteristics of foreground objects and backgrounds are dramatically different, we design a heterogeneous model to handle them separately. To this end, we present \emph{SASPM} and \emph{Boundary-Anchored Patch Discriminator} to facilitate the generation of texture-consistent backgrounds and employ a \emph{Style-Diversity Object Generator} to produce high-quality and diverse objects. Extensive experimental results demonstrate the superiority of our method. However, our method has some limitations. Some foreground categories are very rare in the dataset, so we cannot obtain enough data for training. In addition, object generation is hard to produce satisfactory results when an object is with an extreme pose or large-scale occlusion. 

\noindent
\textbf{Acknowledgement} This work is supported by State Grid Corporation of China (Grant No. 5500-202011091A-0-0-00).

%%%%%%%%% REFERENCES
{\small
\bibliographystyle{ieee_fullname}
\bibliography{egbib}
}

\end{document}